\documentclass[letterpaper,10pt,conference]{ieeeconf}
\IEEEoverridecommandlockouts
\usepackage{graphicx}
\usepackage{amsmath,amssymb}
\usepackage{booktabs}
\usepackage{multirow}
\usepackage{array}
\usepackage{url}
\usepackage{balance}
\usepackage{algorithm}
\usepackage{algorithmic}
\usepackage{hyperref}

\title{\LARGE \bf
CAPABLE: Capability-Aware Policy Adaptation via Behavioral Latent Encoding\\}

\author{%
Mohammad Khoshnazar$^{1,*,\dagger}$,
Mohammad Dehghani Tezerjani$^{2,\dagger}$,
Deyuan Qu$^{3}$,
Zhiyuan Gao$^{1}$,\\
Yanxiang Zhan$^{1}$,
Jeroen Sch\"afer$^{1}$,
Andrew Melnik$^{1}$,
Qing Yang$^{2}$,
Michael Beetz$^{1}$%
\thanks{$^{\dagger}$Mohammad Khoshnazar and Mohammad
Dehghani Tezerjani contributed equally.}%
\thanks{$^{1}$University of Bremen, Bremen, Germany.}%
\thanks{$^{2}$University of North Texas, Denton, TX, USA.}%
\thanks{$^{3}$Toyota Motor North America, InfoTech Labs,
Mountain View, CA, USA.}%
\thanks{$^{*}$Corresponding author.
{\tt\small khoshnazar@uni-bremen.de}}%
}

\usepackage{xcolor}
\definecolor{planblue}{RGB}{0,170,220}
\definecolor{planyellow}{RGB}{220,200,0}
\definecolor{planred}{RGB}{230,30,30}

\begin{document}
\maketitle
\thispagestyle{empty}
\pagestyle{empty}

\begin{abstract}

Vision-language-action (VLA) policies assume the embodiment on which they were trained and can fail when a joint fault changes how commanded actions are physically executed. Existing fault-recovery methods often require task-specific retraining, fault labels, explicit diagnosis, or privileged embodiment information. We introduce CAPABLE, a unified capability-aware adaptation framework for frozen VLAs that integrates self-supervised capability inference with residual reinforcement learning.
CAPABLE infers capability, how much of the commanded motion each joint actually realizes and how that motion contributes to end-effector behavior, online from command--response history and kinematics using a temporal encoder shared across joints, Jacobian grounding, cross-joint attention, and self-supervised physical prediction. The resulting representation conditions a residual policy that adds bounded corrections to the VLA arm action without fault labels or faulty-joint identifiers. Across 28 LIBERO tasks, CAPABLE raises success on an actuator excluded from fault training from 24.8\% to 59.3\%, outperforming a parameter-matched global-history baseline by 17.4 points while preserving healthy performance. Leave-one-actuator-out experiments across six joints show that this transfer is not specific to one actuator, and additional evaluations characterize transfer to unseen fault families and demonstrate recovery on a physical Franka Panda. \href{https://capable-vla.github.io/}{capable-vla.github.io}.
\end{abstract}

\section{Introduction}
A manipulator is midway through placing a plate into a drawer when one
of its motors stalls. The vision-language-action policy driving it is
unaffected. It
continues to issue actions that are semantically correct for the task, and the
arm continues to move, but it no longer moves the way the policy assumes. The
plate never arrives. Incidents of this kind are a reliability and a safety
concern, and are documented in space robotics~\cite{canadarm2002,canadarm2026,curiosity2016}, surgical robotics~\cite{davinci2013}, and industrial manipulators~\cite{flexspline2018,ur5phm2021}.
Prevailing practice is to halt the robot once a fault is detected. Stopping is safe but costly when repair is slow or the robot is hard to reach. We are
interested in the harder alternative, which is to continue the task with a body that has changed.
\begin{figure}[t]
\centering
\includegraphics[width=\columnwidth]{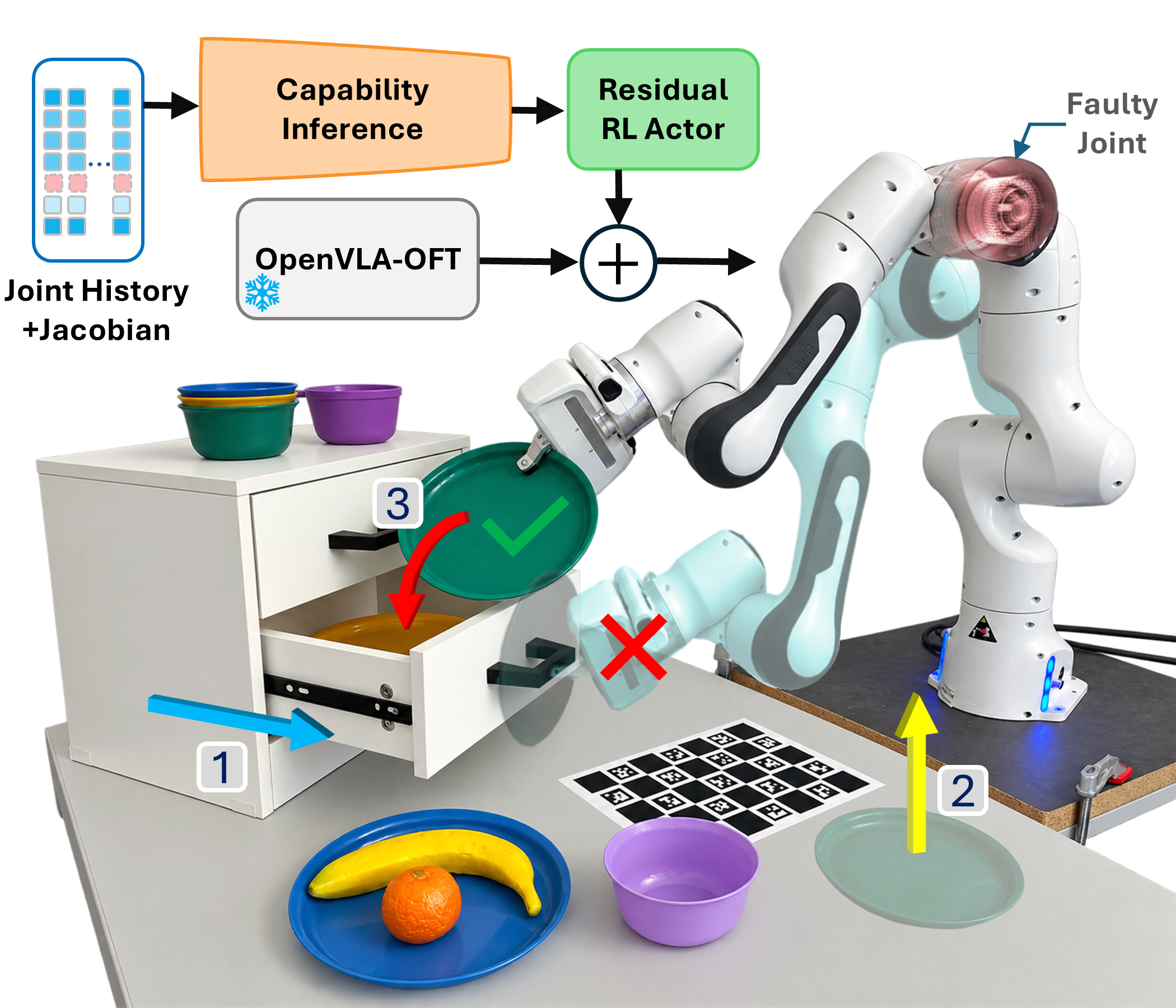}
\caption{The illustrated sequence \textcolor{planblue}{(1) opens the drawer},
\textcolor{planyellow}{(2) lifts the plate}, and
\textcolor{planred}{(3) places it inside}. A faulty joint
(red highlight) prevents the frozen VLA motion from succeeding
(red cross); CAPABLE supplies a corrective motion (green check).
CAPABLE infers each joint's remaining \emph{capability} from its
command--response history and the live Jacobian, and adds a bounded
residual correction to the frozen VLA action. No fault label or explicit
joint-index embedding is supplied, allowing the shared estimator to
transfer to an actuator whose failure was not observed during training.}
\label{fig:teaser}
\end{figure}

Vision-language-action (VLA) policies are an attractive starting point, since
they supply a strong semantic and visuomotor prior for language-conditioned
manipulation~\cite{driess2023palme,brohan2022rt1,zitkovich2023rt2,
ghosh2024octo,black2024pi0,kim2025openvla,kim2025oft}. Their training data,
however, assume the robot retains the healthy embodiment. A joint failure breaks that assumption at the level of physics,
and no amount of visual or linguistic competence repairs it. The fault itself is not directly observable. 
Moreover, the fault space—spanning the affected joint, 
failure type, and severity—is too large to cover during training, requiring 
recovery to generalize to previously unseen actuator failures.

Existing methods address only parts of this problem. Classical fault-tolerant control detects the fault and redistributes motion across the remaining joints~\cite{zhang2008ftc,blanke2006ftc,xie2021locked}, but does not consider the VLA's closed-loop behavior or task progress. Learning-based methods either train a task policy from scratch~\cite{pham2024joint} or require explicit fault information or supervision~\cite{briscoe2026deft,jo2026jparc}. Prior work evaluates held-out joints, but does not establish systematic transfer for a frozen multi-task VLA without fault labels, affected-joint supervision, or healthy reference trajectories.

Our central observation is that a fault need not be identified by its type or cause. Instead, it can be represented by the remaining capability of each actuator, measured by how much of a commanded motion the joint realizes and how that motion affects the end effector in the current configuration. This capability can be inferred without privileged information from each joint’s command–response history. Because capability is represented consistently across all joints, an estimator learned from failures of one joint can transfer to another.

We propose \textbf{CAPABLE (Capability-Aware Policy Adaptation via Behavioral Latent Encoding)}, a residual reinforcement-learning controller that keeps the VLA frozen and adds a correction to its arm action~\cite{johannink2019residual}, using an off-policy actor-critic for
sample-efficient residual policy learning~\cite{ball2023rlpd}.
CAPABLE encodes each joint's command--response history using one temporal encoder shared across all seven joints. The shared map recognizes common command--response signatures across actuators rather than learning a separate encoder per joint (Fig.~\ref{fig:teaser}). Each joint's live Jacobian column grounds this estimate in the current configuration, and cross-joint attention combines all joints into a representation of what motions the arm can still produce. A self-supervised objective that predicts realized joint and end-effector motion ties this representation to physical behavior, and the representation conditions a Soft Actor-Critic policy~\cite{haarnoja2018sac} through FiLM. No fault label, faulty-joint identifier, or fault-specific demonstration enters any network during training or deployment; at deployment the networks are fixed and capability is re-estimated at every control step.

Our contributions are:
\begin{itemize}
\item \textbf{A capability formulation of actuator faults for frozen VLAs}, which represents a fault by what each actuator can still realize, infers it from command--response history and kinematics without fault labels, and uses it to correct a frozen generalist through bounded residual RL.
\item \textbf{A per-actuator factorized architecture for held-out-actuator transfer} that shares one temporal encoder across joints, grounds it in the live Jacobian, composes joints with attention, and trains it with self-supervised physical prediction; removing any of these components significantly reduces held-out-joint success.
\item \textbf{An evaluation that separates factorization from capacity}, comparing against a parameter-matched unfactorized encoder across 28 tasks, six held-out joints, five unseen fault families that mark where transfer holds and where it fails, and a physical robot.
\end{itemize}

\section{Related Work}

\subsection{Adapting Pretrained Robot Policies}
Large robot datasets~\cite{walke2023bridge,khazatsky2024droid} have enabled VLA policies that map visual observations, language instructions, and robot states to actions~\cite{brohan2022rt1,zitkovich2023rt2,ghosh2024octo,black2024pi0,kim2025openvla,kim2025oft}. Because retraining these large models is expensive, recent work adapts them after pretraining. VLA-RL fine-tunes the VLA using reinforcement learning~\cite{lu2025vlarl}, while RobustVLA trains it against observation noise and action perturbations~\cite{zhang2025robustvla}. Other methods use test-time or recovery-oriented reinforcement learning to adapt the policy to new tasks and environments~\cite{liu2026ttvla,liufu2026repo,zhang2026agentic}. Residual learning freezes the base policy and learns correction on the base action~\cite{johannink2019residual,silver2018rpl}. These methods generally assume that the robot's physical capabilities remain unchanged. CAPABLE also keeps the base VLA frozen, but conditions its corrections on an online estimate of the robot's remaining actuator capabilities.

\subsection{Fault-Tolerant Control and Learned Adaptation}

Classical fault-tolerant control follows two main strategies. Passive methods tolerate a predefined set of faults, while active methods diagnose the fault and reconfigure the controller~\cite{zhang2008ftc,blanke2006ftc}. For redundant manipulators, motion can be replanned around a locked joint~\cite{xie2021locked}. Other methods estimate the remaining effectiveness of an actuator before applying compensation~\cite{huang2025estimator}. Both approaches require an explicit model of the fault. Learning-based methods instead infer changes in the robot's dynamics from recent interaction history~\cite{yu2017uposi,kumar2021rma,nagabandi2019meta}. Fault-tolerant locomotion policies can recover from locked or unpowered joints, but their training often uses privileged fault information through teacher policies, privileged critics, or fault-specific experts~\cite{liu2022limping,xu2025acl,gravina2026gait,turrisi2026moe}.

\subsection{Learning-Based Fault Recovery}
Pham et al.~\cite{pham2024joint} formulate Franka joint failure as a POMDP and
train a recurrent PPO policy for permanent and intermittent failures on a
drawer task, establishing that history-based~\cite{tezerjani2024slam,khoshnazar2026future} can compensate for an unknown
impaired actuator. Their policy is trained for the task directly rather than
wrapped around a pretrained multi-task VLA. DEFT~\cite{briscoe2026deft} uses an
embodiment- and task-conditioned diffusion model to generate fail-active
trajectories with broad simulated failure coverage, conditioning on an explicit
embodiment vector. J-PARC~\cite{jo2026jparc} is the closest VLA-specific study. 
It trains an action calibrator on locked-joint faults using healthy reference 
rollouts and fault-presence and affected-joint supervision, then evaluates transfer 
to restricted ranges, friction, and held-out joints. CAPABLE uses task reward and self-supervised physical
prediction in place of all of these signals.

\subsection{Weight Sharing Across Actuators}
Sharing one module across the parts of a body is a well-established route to
morphology-agnostic control. NerveNet propagates messages over the kinematic
graph~\cite{wang2018nervenet}, shared modular policies apply a single policy to
every limb~\cite{huang2020smp}, and transformer variants attend over limb tokens~\cite{kurin2021amorpheus,
gupta2022metamorph}. That literature shares modules across \emph{morphologies} to generalize over
bodies. We share a module across \emph{actuators of one body} to generalize
over which actuator has failed, grounding the shared tokens in the live
Jacobian so the estimate becomes a configuration-specific correction. The components themselves are standard,
combining recurrent state for partial
observability~\cite{hausknecht2015drqn,cho2014gru}, attention for composition
across entities~\cite{vaswani2017attention}, and FiLM for
conditioning~\cite{perez2018film}.

\section{Problem Formulation}

Let the pretrained VLA $\pi_{\phi}$ receive observation $o_t$ and language
instruction $\ell$ and produce a seven-dimensional action
\begin{equation}
 a_t^{\mathrm{VLA}}=\pi_{\phi}(o_t,\ell)
 = [a_{\mathrm{arm},t}^{\mathrm{VLA}},
    a_{\mathrm{grip},t}^{\mathrm{VLA}}].
\end{equation}
The VLA parameters $\phi$ remain fixed during residual training.

A persistent fault \(f\) changes the nominal transition dynamics \(p_0(s_{t+1}\mid s_t,a_t)\) to fault-conditioned dynamics \(p_f(s_{t+1}\mid s_t,a_t)\). The policy does not observe \(f\) and must therefore infer its physical effects from execution history.

 The actor produces
\begin{equation}
 u_t=\pi_{\theta}(\widetilde{o}_t^{\mathrm{RL}},z_t^{\mathrm{cap}}),
 \qquad u_t\in[-1,1]^6,
\end{equation}
and the applied correction is
\begin{equation}
 \Delta a_t=\eta u_t,\qquad \eta=0.1.
\end{equation}
The executed action is
\begin{equation}
 a_t^{\mathrm{exec}}=
 \left[
 \operatorname{clip}\!\left(
 a_{\mathrm{arm},t}^{\mathrm{VLA}}+\Delta a_t,-1,1
 \right),
 a_{\mathrm{grip},t}^{\mathrm{VLA}}
 \right].
 \label{eq:correction}
\end{equation}
The gripper passes through untouched; the sparse reward is the LIBERO
task-success predicate.
An \emph{unseen actuator} is a joint whose failure is excluded from 
all stages of training and encountered only during evaluation.

\section{CAPABLE: Capability-Aware Policy Adaptation via Behavioral Latent Encoding}

\begin{figure*}[t]
\centering
\includegraphics[width=0.99\textwidth]{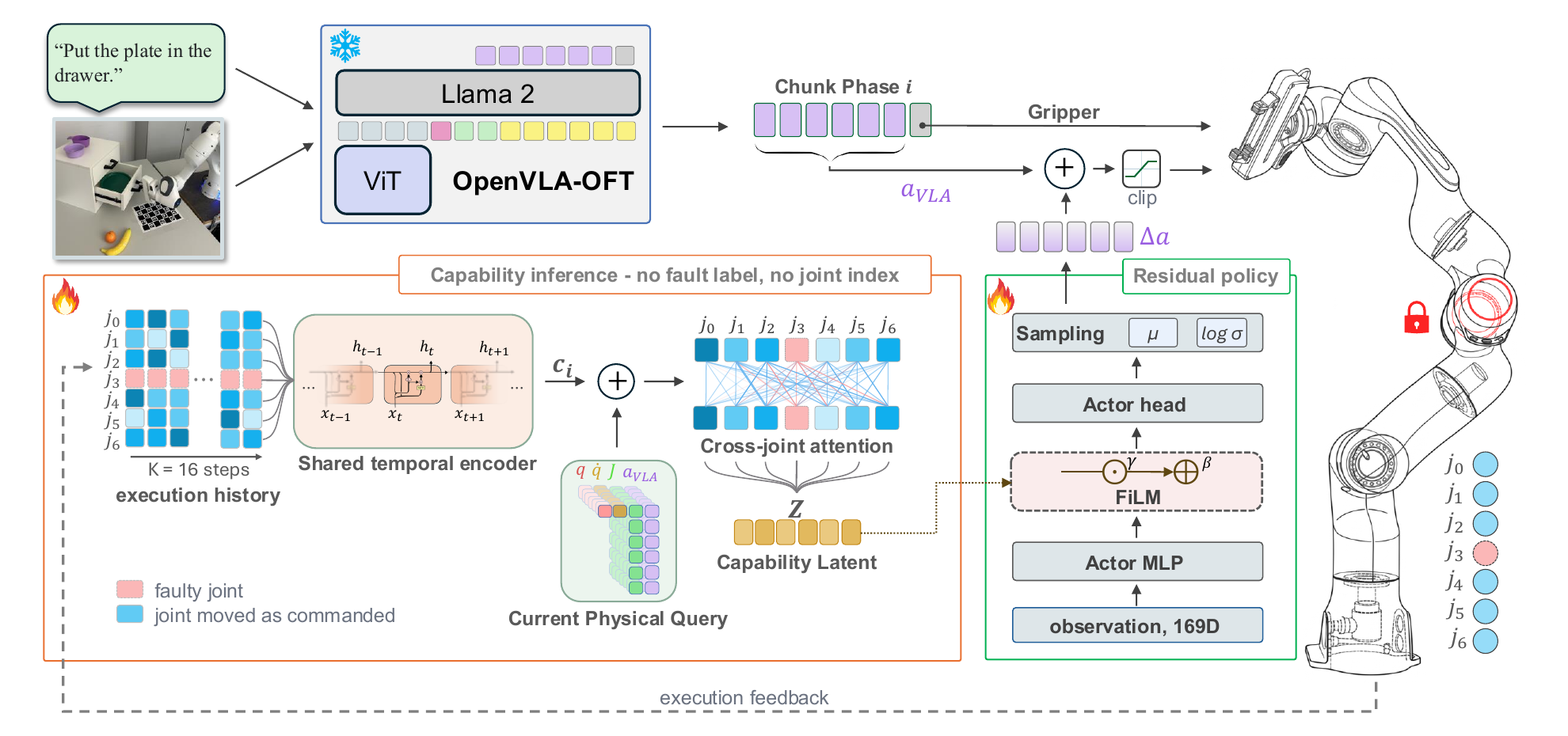}
\caption{CAPABLE architecture. Shared encoders map per-joint command--response
histories and live kinematics to $z_t^{\mathrm{cap}}$, which FiLM-conditions
the residual SAC actor. The bounded arm correction is added to the frozen VLA
action; the gripper bypasses RL. Critics, replay and rewards are training-only.}
\label{fig:architecture}
\end{figure*}

CAPABLE combines shared per-joint history encoding, current kinematic
grounding, and self-supervised physical prediction to infer actuator
capability and generate residual corrections. Figure~\ref{fig:architecture}
shows the architecture, and Algorithm~\ref{alg:jfcrl} gives deployment.

\subsection{Per-Joint Command--Response History}
\label{sec:token}

For Panda arm joint $i\in\{0,\ldots,6\}$, CAPABLE constructs the token
\begin{align}
 x_t^i = [&q_t^i,\dot q_t^i,\Delta q_t^i,J_i(q_t),
 a_{\mathrm{arm},t}^{\mathrm{VLA}},a_{\mathrm{arm},t}^{\mathrm{exec}},
 \nonumber\\
 &\Delta x_t^{\mathrm{eef}},\rho_t],
 \label{eq:jointtoken}
\end{align}
where $J_i(q_t)\in\mathbb{R}^{6}$ is the $i$th column of the $6\times7$ arm
Jacobian, $\Delta x_t^{\mathrm{eef}}\in\mathbb{R}^{6}$ is realized
end-effector change, and $\rho_t$ is the VLA action-chunk phase. Each token is
28-dimensional vector with the same feature structure across all joints. Identity enters the
token geometrically, through $J_i(q_t)$, which describes how motion of joint \(i\) contributes to end-effector motion in the current configuration. 
The model uses no one-hot joint indicator or learned joint-index embedding, 
and no lock target, fault-family label, or severity value is provided to the encoder, actor, or critics during training or deployment.
\subsection{Shared Temporal Encoder}
\label{sec:sharedenc}

Each joint stores $K=16$ recent tokens. One encoder with identical weights for
all joints produces
\begin{equation}
 c_t^i=E_{\psi}(x_{t-K:t-1}^i),
 \qquad c_t^i\in\mathbb{R}^{32}.
\end{equation}
The history contains past information while the current-step information
enters only through the physical query. $E_{\psi}$ contains a linear projection, LayerNorm, SiLU,
and a GRU with hidden size 128. Weight sharing is what makes capability
transferable. Because the encoder weights are shared, a command–response pattern learned 
on one joint can be recognized on any other joint, including one whose failure was never observed during training.

\subsection{Kinematic Query and Cross-Joint Composition}
The current query for joint $i$ is
\begin{equation}
 \kappa_t^i=[q_t^i,\dot q_t^i,J_i(q_t),
 a_{\mathrm{arm},t}^{\mathrm{VLA}}].
 \label{eq:query}
\end{equation}
For each joint \(i\), we concatenate its history embedding \(c_t^i\) with its current kinematic query \(\kappa_t^i\) and linearly project the result to form a joint token. A two-layer, four-head Transformer encoder with feed-forward width 256 then applies self-attention across the seven joint tokens, allowing information to be exchanged among joints. No joint-ID embedding is used. Finally, we average the seven output tokens and pass the result through an
MLP and LayerNorm to obtain $z_t^{\mathrm{cap}}\in\mathbb{R}^{64}$.
The command--response history shows how much each joint actually moves when commanded. 
The Jacobian maps this joint motion to its effect on the end effector. Combining the estimates 
from all seven joints describes what motions the impaired arm can still produce.

\subsection{Self-Supervised Physical Prediction}
\label{sec:aux}
A shared local decoder predicts realized joint displacement and a global
decoder predicts realized end-effector displacement:
\begin{align}
\widehat{\Delta q}_t^i
&=D_{\mathrm{joint}}(c_t^i,\kappa_t^i,\Delta a_t),\\
 \widehat{\Delta x}_t^{\mathrm{eef}}
 &=D_{\mathrm{eef}}(z_t^{\mathrm{cap}},o_t^{\mathrm{RL}},\Delta a_t).
\end{align}
Here $\widehat{\Delta q}_t=
[\widehat{\Delta q}_t^0,\ldots,\widehat{\Delta q}_t^6]^\top
\in\mathbb{R}^{7}$ stacks the per-joint predictions from the shared local
decoder. Using Huber loss $\ell_H$, the objectives are
\begin{align}
 \mathcal{L}_{\mathrm{joint}}
 &=\frac{1}{7}\sum_i
 \ell_H(\widehat{\Delta q}_t^i,\Delta q_t^i),\\
 \mathcal{L}_{\mathrm{eef}}
 &=\ell_H(\widehat{\Delta x}_t^{\mathrm{eef}},
 \Delta x_t^{\mathrm{eef}}),\\
 \mathcal{L}_{\mathrm{kin}}
 &=\ell_H(J(q_t)\widehat{\Delta q}_t,
 \Delta x_t^{\mathrm{eef}}),\\
 \mathcal{L}_{\mathrm{cap}}
 &=\mathcal{L}_{\mathrm{joint}}+\mathcal{L}_{\mathrm{eef}}
 +0.25\mathcal{L}_{\mathrm{kin}}.
 \label{eq:caploss}
\end{align}
\(\mathcal{L}_{\mathrm{kin}}\) enforces kinematic consistency by requiring the predicted joint 
displacements, mapped through the Jacobian, to match the observed end-effector displacement.

\subsection{FiLM-Conditioned SAC and Gradient Routing}
\label{sec:sac}

CAPABLE uses twin-critic SAC~\cite{haarnoja2018sac}. The 169-dimensional residual
observation is
\begin{equation}
\begin{aligned}
o_t^{\mathrm{RL}}=[&
p_t^{\mathrm{eef}},r_t^{\mathrm{eef}},q_t^{\mathrm{grip}},
q_t,\dot q_t,a_t^{\mathrm{VLA}},\rho_t,\tau_t,\\
&h_{t-8:t-1},
\operatorname{vec}_{\mathrm{joint}}(J(q_t))],
\end{aligned}
\label{eq:rlobs}
\end{equation}
where $p_t^{\mathrm{eef}},r_t^{\mathrm{eef}}\in\mathbb{R}^{3}$ are measured
end-effector position and orientation,
$q_t^{\mathrm{grip}}\in\mathbb{R}^{2}$ the gripper joints,
$q_t,\dot q_t\in\mathbb{R}^{7}$ the measured arm state, $\rho_t$ the
action-chunk phase, and $\tau_t=t/T_{\max}$ normalized episode time. Each
execution-history element is
\begin{equation}
h_k=[a_{\mathrm{arm},k}^{\mathrm{exec}},
\Delta p_k^{\mathrm{eef}},\Delta r_k^{\mathrm{eef}}]
\in\mathbb{R}^{12},
\end{equation}
All input components are obtained from 
robot sensing, controller outputs, internal counters, or the known kinematic model.
 The actor and critics receive the normalized observation
$\widetilde{o}_t^{\mathrm{RL}}=(o_t^{\mathrm{RL}}-\mu_o)/(\sigma_o+\epsilon)$,
and FiLM modulates their intermediate features $v_t$:
\begin{align}
 \gamma_t&=1+0.5\tanh(\gamma_{\mathrm{raw}}(z_t^{\mathrm{cap}})),\\
 \beta_t&=0.5\tanh(\beta_{\mathrm{raw}}(z_t^{\mathrm{cap}})),\\
 v_t'&=\gamma_t\odot v_t+\beta_t.
\end{align}
The actor's deterministic output is initialized to zero, so training begins
from exact reproduction of the base VLA arm action.

The capability encoder is optimized by the auxiliary physical-prediction loss and a critic loss weighted by \(0.05\):
\begin{equation}
\nabla_{\psi}\mathcal{L}
=
\nabla_{\psi}\mathcal{L}_{\mathrm{cap}}
+
0.05\,\nabla_{\psi}\mathcal{L}_{Q}.
\end{equation}
Actor-loss gradients are blocked at \(z_t^{\mathrm{cap}}\) and therefore do not update \(E_{\psi}\), keeping encoder learning dominated by the physical-prediction objective.

\begin{algorithm}[t]
\caption{CAPABLE residual control (deployment)}
\label{alg:jfcrl}
\begin{algorithmic}[1]
\REQUIRE frozen VLA $\pi_\phi$, encoder $E_\psi$, actor $\pi_\theta$,
chunk length $L=8$, scale $\eta=0.1$
\STATE initialize per-joint buffers $\mathcal{X}^i\!\leftarrow\!\emptyset$,
$i=0..6$
\FOR{control step $t=0,1,\ldots$}
  \IF{$t \bmod L = 0$}
    \STATE $a^{\mathrm{VLA}}_{t:t+L-1}\leftarrow\pi_\phi(o_t,\ell)$
    \hfill\COMMENT{new chunk}
  \ENDIF
  \STATE $\rho_t\leftarrow (t \bmod L)/L$
  \STATE $c_t^i\leftarrow E_\psi(\mathcal{X}^i[t\!-\!K:t\!-\!1])\;\forall i$
  \hfill\COMMENT{shared weights}
  \STATE $\kappa_t^i\leftarrow[q_t^i,\dot q_t^i,J_i(q_t),
a^{\mathrm{VLA}}_{\mathrm{arm},t}]\;\forall i$
  \STATE $z_t^{\mathrm{cap}}\leftarrow
  \mathrm{Attn}(\{c_t^i,\kappa_t^i\}_{i=0}^{6})$
  \STATE $u_t\leftarrow\pi_\theta(\widetilde o_t^{\mathrm{RL}},
  z_t^{\mathrm{cap}})$; \quad $\Delta a_t\leftarrow\eta u_t$
  \STATE execute $a_t^{\mathrm{exec}}$ by Eq.~\eqref{eq:correction}
  \STATE observe $\Delta q_t,\Delta x_t^{\mathrm{eef}}$;
  append $x_t^i$ to $\mathcal{X}^i$
\ENDFOR
\end{algorithmic}
\end{algorithm}

\section{Experimental Setup}

\subsection{Benchmark, Simulator, and Base Policy}

We use LIBERO~\cite{liu2023libero} with the Franka Panda implementation in
robosuite~\cite{zhu2020robosuite} and MuJoCo~\cite{todorov2012mujoco}. The base
checkpoint is
\texttt{moojink/openvla-7b-oft-}\allowbreak
\texttt{finetuned-libero-spatial-}\allowbreak
\texttt{object-goal-10},
loaded once and shared across all tasks, with suite-specific action statistics
used for unnormalization. OpenVLA-OFT executes action chunks of length eight.
The residual policy changes only the six continuous arm dimensions.

A joint lock is a MuJoCo equality constraint on the physical Panda joint rather
than an action mask, so the arm remains dynamically coupled to the locked link.
Condition changes rebuild the relevant environment path, and lock drift is
monitored. The 28-task and cross-fault studies are conducted in simulation;
Sec.~\ref{sec:realrobot} evaluates the trained residual policy on a physical
Franka Panda under software-enforced joint locks.

\subsection{Held-Out-Blind 28-Task Pool}
\label{sec:curriculum}

We screen all 40 LIBERO tasks with the base VLA, using only healthy outcomes
and candidate training conditions
$\mathcal{J}_{\mathrm{seen}}=\{j_0,j_4,j_5,j_6\}$. A task is retained when healthy success is at least 80\% 
and at least one seen fault is \emph{usable}, meaning that the corresponding fault condition has 
nonzero success and sufficient headroom for improvement. Twelve tasks fail this rule, leaving 28 tasks and 46
usable task--fault cells, comprising ten from LIBERO-10, seven from
LIBERO-Spatial, six from LIBERO-Object, and five from LIBERO-Goal. The selected task indices are
supplementary.

Failures on $j_2$ are excluded from all training stages.
Its screening results are used only for reporting, so all 16 tasks with 0\%
base-VLA success under $j_2$ remain valid evaluation cases.

\subsection{Task Curriculum and Replay Distribution}
Tasks are sampled uniformly, with $p(m)=1/28$. Within each task, healthy execution
is sampled with probability 0.10, while usable seen faults share the remaining 0.90
in proportion to their screened headroom $s_m^{\mathrm{H}}-s_{m,j}$, where
$s_m^{\mathrm{H}}$ is healthy base-VLA success for task $m$ and $s_{m,j}$ is
success under locked joint $j$.

Replay uses the same task--condition distribution. Each minibatch weights tasks
equally and samples conditions according to the rule above.
The pool contains 28 healthy task--condition pairs among 74 total pairs, so
uniform sampling over all pairs would allocate $28/74\approx37.8\%$ of updates
to healthy data instead of the target 10\%. Section~\ref{sec:ablations} evaluates this alternative.

\subsection{Training and Deterministic Evaluation}
For each task, a held-out set of initial states is reserved before any
learning and is disjoint from the states available to training. Every method is
trained for 1.4M environment steps per seed, and evaluated at the 1.4M-step checkpoint. 
We run three independently initialized seeds sharing the
task pool, state partition, data budget, base VLA, and evaluation procedure;
reported $\pm$ values are mean $\pm$ standard deviation across these seeds.

Evaluation compares zero correction with each learned policy on healthy
execution, every usable seen fault, and globally unseen $j_2$. Policy
actions are deterministic, and within a seed each task--condition cell uses the
same 300 held-out episodes for all methods. We additionally perform leave-one-actuator-out evaluation for six joints
($j_0$--$j_5$) on eight tasks, with two tasks selected from each LIBERO suite.
In each split, the held-out joint is excluded from fault training and evaluated
zero-shot.

We report task-balanced success, averaging seen-fault success within task
before averaging across tasks, so tasks with more usable faults carry no extra
weight.

All architecture, optimization, replay, and fault-severity hyperparameters
were fixed using training data and validation or pilot episodes involving only
seen joints. Held-out $j_2$ outcomes were never used for task selection,
checkpoint selection, hyperparameter tuning, or other design decisions.

\subsection{Cross-Fault Generalization Protocol}
\label{sec:crossfault_protocol}

The headline CAPABLE and global-history policies are trained only on persistent
locks affecting $j_0,j_4,j_5,$ and $j_6$. Without retraining, we evaluate the same three seed checkpoints
on globally unseen $j_2$ under six fault families, namely a persistent lock, increased
viscous damping, increased Coulomb friction, partial actuator effectiveness,
restricted joint range, and a late-onset lock.

The five unseen fault families are evaluated at three levels: viscous joint
damping $d\in\{5,20,50\}\,\mathrm{N\,m\,s\,rad^{-1}}$, Coulomb friction
$f\in\{1,3,5\}\,\mathrm{N\,m}$, retained actuator authority and joint-range
fractions in $\{0.75,0.50,0.25\}$, and lock onset at control step
$t_{\mathrm{on}}\in\{32,64,96\}$.
Severity levels are fixed from seen-joint pilots without using $j_2$ outcomes.
All methods use the same 300 held-out episodes per task--condition cell. For
multi-severity faults, success is averaged over severities within each task and
seed, then equally across the 28 tasks.

\subsection{Baselines}

The central baseline is \textbf{Global-history SAC}, which isolates the
factorization prior. It shares the VLA, residual action space, 16-step window,
64-D latent, FiLM actor and twin critics, auxiliary targets, optimizer, replay
distribution, budget, and evaluation states with CAPABLE, but concatenates the
seven 28-D joint tokens at each step and encodes them with one global recurrent
model, with no shared per-joint encoder and no cross-joint token structure. Hidden width is adjusted
so trainable parameter count is within 1\% of CAPABLE. Both models use SAC hidden
width 256, encoder and SAC learning rates $10^{-4}$ and $3\times10^{-4}$, batch
size 256, UTD 4, and $n$-step 3.
\textbf{Classical RR} is a non-learning redundancy-resolution baseline.
The same unnormalized six-dimensional VLA arm command is used as the desired
Cartesian motion. Joint availability is estimated online from recent
command--response mismatch, and a weighted damped pseudoinverse of the live
Jacobian redistributes the command toward responsive joints. It receives no
fault label or affected-joint identifier, and the gripper passes through
unchanged.

\textbf{POMDP-RL}~\cite{pham2024joint} and \textbf{DEFT}~\cite{briscoe2026deft}
are trained in their own formulations on the same 28-task pool, fault
conditions, training budget, and held-out evaluation states. POMDP-RL learns a
recurrent task policy under partial observability. DEFT generates joint-space
trajectories between a start and a goal configuration, so it receives goal
configurations derived from simulator object and target-region state through
inverse kinematics, a scripted gripper at phase boundaries, and the current
actuator-availability vector. It is therefore privileged in two respects, on
task state and on fault identity, whereas CAPABLE uses only non-privileged
runtime observations, controller signals, and known kinematics; its result measures
the value of that privileged information rather than providing a matched comparison. The matched-information comparison
for the factorization claim is CAPABLE versus Global-history SAC.
Table~\ref{tab:baseline_assumptions} summarizes each method's assumptions. We excluded J-PARC~\cite{jo2026jparc} as a similar baseline due to the absence of peer-reviewed validation and publicly released implementation.

\begin{table}[t]
\caption{Comparison assumptions. Learned methods use the same task pool,
fault conditions, budget, and held-out states. Classical RR is non-learning;
DEFT receives simulator-derived goals and actuator availability.}
\label{tab:baseline_assumptions}
\centering
\scriptsize
\setlength{\tabcolsep}{3pt}
\resizebox{\columnwidth}{!}{%
\begin{tabular}{@{}lccccc@{}}
\toprule
Method & Base VLA & Residual & Oracle fault & Oracle task state
& Joint-fact. \\
\midrule
Base VLA (no correction) & Yes & No  & No & No & No \\
POMDP-RL & No & No & No & No & No \\
Classical RR & Yes & No & No & No & \textbf{Yes} \\
DEFT & No & No & \textbf{Yes} & \textbf{Yes} & No \\
Global-history SAC & Yes & Yes & No & No & No \\
CAPABLE (ours) & Yes & Yes & No & No & \textbf{Yes} \\
\bottomrule
\end{tabular}%
}
\end{table}

\section{Results}
\label{sec:results}

We organize the evaluation around the three claims of Sec.~I.
\textbf{H1:} does joint factorization transfer capability to an actuator that
was never faulty during training, beyond what an unfactorized encoder of equal
capacity achieves?
\textbf{H2:} does capability learned from persistent locks transfer to fault
families never seen in training?
\textbf{H3:} does held-out-actuator transfer generalize systematically
across actuator identity, and does it survive contact with hardware?

\subsection{H1: Factorization Transfers to a Held-Out Actuator}

Table~\ref{tab:main_results} is the primary result. Across 28 tasks, CAPABLE
raises unseen-$j_2$ success from $24.8\%$ to $59.3\%$, versus $41.9\%$ for
Global-history SAC and $32.9\%$ for classical redundancy resolution. The paired CAPABLE--global gap is $+17.4$ points under matched capacity,
data, optimization, and evaluation. CAPABLE also exceeds classical redundancy
resolution by $26.4$ points on unseen $j_2$. Healthy success remains $90.8\%$
versus $91.4\%$ for zero correction. DEFT reaches $62.8\%$ with oracle fault
and task-state information, only 3.5 points above CAPABLE.

\begin{table}[!ht]
\caption{Main 28-task success (\%). $H/S/U$: healthy/seen/unseen-$j_2$;
$\Delta$ is versus Base VLA on unseen $j_2$. Classical RR is non-learning;
DEFT uses oracle fault and task state.}
\label{tab:main_results}
\centering
\scriptsize
\setlength{\tabcolsep}{5pt}
\resizebox{\columnwidth}{!}{%
\begin{tabular}{@{}lcccc@{}}
\toprule
Method & $H$ & $S$ & $U$ & $\Delta$ vs. Base \\
\midrule
Base VLA
& $\mathbf{91.4\!\pm\!0.0}$
& $41.6\!\pm\!0.0$
& $24.8\!\pm\!0.0$
& -- \\

POMDP-RL
& $19.6\!\pm\!2.4$
& $11.3\!\pm\!2.1$
& $10.1\!\pm\!1.9$
& $-14.7$ \\

Classical RR
& $81.6\!\pm\!0.0$
& $61.3\!\pm\!0.0$
& $32.9\!\pm\!0.0$
& $+8.1$ \\

DEFT
& $89.7\!\pm\!1.4$
& $70.1\!\pm\!2.3$
& $\mathbf{62.8\!\pm\!2.9}$
& $\mathbf{+38.0}$ \\

Global-history SAC
& $88.1\!\pm\!1.3$
& $73.8\!\pm\!2.1$
& $41.9\!\pm\!3.2$
& $+17.1$ \\

\textbf{CAPABLE}
& $90.8\!\pm\!1.0$
& $\mathbf{86.7\!\pm\!1.8}$
& $59.3\!\pm\!3.0$
& $+34.5$ \\
\bottomrule
\end{tabular}%
}
\end{table}

Table~\ref{tab:suite_results} shows the advantage is not suite-specific:
CAPABLE leads Global-history SAC on all four suites and 24 of 28 task cells,
with margins from $+12.8$ to $+24.3$ points. Zero-baseline cells are retained
in all averages; full task-level results are supplementary.

\begin{table}[!ht]
\caption{Unseen-$j_2$ success by suite (\%).}
\label{tab:suite_results}
\centering
\scriptsize
\setlength{\tabcolsep}{5pt}
\resizebox{\columnwidth}{!}{%
\begin{tabular}{lcccc}
\toprule
Suite & Base VLA & Global-history SAC & CAPABLE & $\Delta$ vs. Global \\
\midrule
LIBERO-10
& $8.0\!\pm\!0.0$ & $26.5\!\pm\!4.6$ & $\mathbf{45.2\!\pm\!4.1}$
& $+18.7$ \\

LIBERO-Goal
& $40.0\!\pm\!0.0$ & $49.0\!\pm\!3.8$ & $\mathbf{73.3\!\pm\!4.7}$
& $+24.3$ \\

LIBERO-Object
& $20.8\!\pm\!0.0$ & $52.8\!\pm\!3.3$ & $\mathbf{65.6\!\pm\!4.4}$
& $+12.8$ \\

LIBERO-Spatial
& $41.4\!\pm\!0.0$ & $49.5\!\pm\!1.2$ & $\mathbf{64.0\!\pm\!3.8}$
& $+14.5$ \\
\midrule
Task-balanced mean
& $24.8\!\pm\!0.0$ & $41.9\!\pm\!3.2$ & $\mathbf{59.3\!\pm\!3.0}$
& $+17.4$ \\
\bottomrule
\end{tabular}%
}
\end{table}

\subsection{H2: Transfer Extends to Unseen Fault Families}

Table~\ref{tab:crossfault} tests zero-shot transfer on unseen $j_2$ across the
trained persistent lock and five unseen fault families. CAPABLE leads
Global-history SAC on persistent lock, damping, friction, and late-onset lock,
but not on partial effectiveness or range restriction. The successful cases
share the trained signature of suppressed motion relative to command, whereas
the latter two are graded or state-dependent.

\begin{table*}[t]
\caption{Zero-shot cross-fault success on unseen $j_2$ (\%).
Training uses only persistent locks; non-lock columns average three severities.}
\label{tab:crossfault}
\centering
\scriptsize
\setlength{\tabcolsep}{7pt}
\begin{tabular}{lcccccc}
\toprule
Method & Persistent lock & Damping & Friction & Partial effectiveness
& Range restriction & Late-onset lock \\
\midrule
Base VLA
& $24.8\!\pm\!0.0$ & $45.8\!\pm\!0.0$ & $40.6\!\pm\!0.0$
& $44.2\!\pm\!0.0$ & $32.7\!\pm\!0.0$ & $50.4\!\pm\!0.0$ \\

Global-history SAC
& $41.9\!\pm\!3.2$ & $55.6\!\pm\!1.4$ & $51.8\!\pm\!1.6$
& $\mathbf{64.1\!\pm\!3.3}$ & $\mathbf{54.7\!\pm\!2.8}$ & $61.5\!\pm\!3.1$ \\

\textbf{CAPABLE}
& $\mathbf{59.3\!\pm\!3.0}$ & $\mathbf{67.8\!\pm\!2.9}$
& $\mathbf{64.2\!\pm\!3.2}$ & $62.3\!\pm\!3.0$ & $48.2\!\pm\!3.5$
& $\mathbf{72.4\!\pm\!2.7}$ \\
\midrule
$\Delta$ vs. Global
& $+17.4$ & $+12.2$
& $+12.4$ & $-1.8$
& $-6.5$ & $+10.9$ \\
\bottomrule
\end{tabular}
\end{table*}

\subsection{H3: Transfer Generalizes Across Held-Out Actuators}

Table~\ref{tab:reverse_split} tests actuator-independent transfer across six
held-out joints on the same eight LIBERO tasks. The $j_2$ entry uses this
common eight-task subset, whereas Table~\ref{tab:main_results} reports the
primary 28-task $j_2$ result. CAPABLE outperforms Global-history SAC on all six
held-out actuators, showing that its transfer advantage is not specific to
$j_2$.

\begin{table}[!ht]
\caption{Leave-one-actuator-out transfer across six held-out joints on eight
LIBERO tasks. Each row uses a separate leave-one-actuator-out training split in which the
indicated joint is excluded from fault training and evaluated zero-shot. Values are success rates (\%).}
\label{tab:reverse_split}
\centering
\scriptsize
\setlength{\tabcolsep}{4pt}
\resizebox{\columnwidth}{!}{%
\begin{tabular}{lccc}
\toprule
Held-out joint & Global-history SAC & CAPABLE & $\Delta$ vs. Global \\
\midrule
$j_0$ & $55.0\!\pm\!4.4$ & $\mathbf{76.7\!\pm\!3.3}$ & $+21.7$ \\
$j_1$ & $39.4\!\pm\!1.3$ & $\mathbf{56.2\!\pm\!2.3}$ & $+16.8$ \\
$j_2$ & $46.6\!\pm\!5.2$ & $\mathbf{69.3\!\pm\!3.0}$ & $+22.7$ \\
$j_3$ & $39.6\!\pm\!4.6$ & $\mathbf{63.6\!\pm\!4.7}$ & $+24.0$ \\
$j_4$ & $44.7\!\pm\!5.1$ & $\mathbf{88.1\!\pm\!1.9}$ & $+43.4$ \\
$j_5$ & $29.3\!\pm\!4.2$ & $\mathbf{48.2\!\pm\!3.6}$ & $+18.9$ \\
\midrule
Mean & $42.4$ & $\mathbf{67.0}$ & $+24.6$ \\
\bottomrule
\end{tabular}%
}
\end{table}

\subsection{Mechanism Ablations}
\label{sec:ablations}

Table~\ref{tab:ablations} isolates which components produce the transfer.
Removing history leaves a memoryless residual with current Jacobian information
at 39.4\% on unseen $j_2$, below Global-history SAC, so static kinematic
redirection does not explain the result. Unsharing the temporal encoder costs
13.5 points on unseen $j_2$ and 6.6 on seen faults; removing the Jacobian query
costs 11.5 points, cross-joint attention 13.2, and the self-supervised capability
loss 15.6, one of the largest drops. Removing $z^{\mathrm{cap}}$ conditioning
costs 9.7. Together these ablations support a coupled temporal--kinematic
representation shaped by physical prediction rather than any single component.
The unshared variant also performs worse with 50\% more parameters.
Equal-strata replay raises healthy success to 92.3\% while reducing fault
recovery, supporting the default headroom-weighted replay distribution.

\begin{table}[!ht]
\caption{28-task ablations. $P$: parameters (M); $H/S/U$:
healthy/seen/unseen-$j_2$; $\Delta$ is versus full CAPABLE on unseen $j_2$.}
\label{tab:ablations}
\centering
\scriptsize
\setlength{\tabcolsep}{3pt}
\resizebox{\columnwidth}{!}{%
\begin{tabular}{@{}lccccc@{}}
\toprule
Variant & $P$ & $H$ & $S$ & $U$ & $\Delta$ vs. Full \\
\midrule
\textbf{Full CAPABLE}
& $0.82$ & $90.8\!\pm\!1.0$ & $\mathbf{86.7\!\pm\!1.8}$
& $\mathbf{59.3\!\pm\!3.0}$ & -- \\

Global-history SAC
& $0.82$ & $88.1\!\pm\!1.3$ & $73.8\!\pm\!2.1$ & $41.9\!\pm\!3.2$
& $-17.4$ \\

Equal-strata replay
& $0.82$ & $\mathbf{92.3\!\pm\!0.9}$ & $71.5\!\pm\!2.6$ & $46.4\!\pm\!3.7$
& $-12.9$ \\

No Jacobian query
& $0.81$ & $89.8\!\pm\!1.4$ & $75.9\!\pm\!2.4$ & $47.8\!\pm\!3.8$
& $-11.5$ \\

No cross-joint attention
& $0.71$ & $89.2\!\pm\!1.5$ & $81.2\!\pm\!2.7$ & $46.1\!\pm\!3.6$
& $-13.2$ \\

No capability loss
& $0.80$ & $88.7\!\pm\!1.7$ & $72.0\!\pm\!3.1$ & $43.7\!\pm\!4.6$
& $-15.6$ \\

Concatenation, not FiLM
& $0.84$ & $90.1\!\pm\!1.3$ & $78.4\!\pm\!2.3$ & $52.6\!\pm\!3.4$
& $-6.7$ \\

No $z^{\mathrm{cap}}$ conditioning
& $0.76$ & $89.7\!\pm\!1.6$ & $76.6\!\pm\!2.7$ & $49.6\!\pm\!3.9$
& $-9.7$ \\

Unshared per-joint encoders
& $1.23$ & $90.9\!\pm\!1.1$ & $80.1\!\pm\!2.3$ & $45.8\!\pm\!4.2$
& $-13.5$ \\

No-history residual
& $0.60$ & $90.6\!\pm\!1.4$ & $68.4\!\pm\!3.1$ & $39.4\!\pm\!4.4$
& $-19.9$ \\
\bottomrule
\end{tabular}%
}
\end{table}

\subsection{Hardware Validation}
\label{sec:realrobot}

We evaluate CAPABLE and parameter-matched Global-history SAC on a physical Franka
Panda under three task--joint conditions, opening a drawer with seen $j_0$,
putting the object in the drawer with unseen $j_2$, and object rearrangement with
seen $j_6$. Each trial carries one persistent lock, software-enforced at the
joint-reference layer, so that once end-effector commands are converted to
joint references, the selected joint is clamped to its fault-onset angle while all
others remain controllable. Commanding the held angle rather than mechanically
opposing motion avoids triggering the Franka reflex path.

For hardware transfer, OpenVLA-OFT is initialized from
the Sec. V-A checkpoint,
fine-tuned in a digital twin matched to the physical workspace, and then frozen.
CAPABLE and Global-history SAC are subsequently trained on top of this frozen VLA
in the same simulated setup and transferred to the physical Panda without
hardware training. The digital twin matches the robot/table geometry, camera
pose, objects, controller, and high-level control rate, with randomized object
poses and physical parameters. Each method receives 10 trials per condition.

CAPABLE reaches 85.0\% across the two seen-joint conditions against
45.0\% for Global-history SAC, and 70.0\% against 40.0\% on the
unseen-$j_2$ condition, for an overall 80.0\% versus 43.3\% across
30 trials per method (Table~\ref{tab:real_robot}).
Individual conditions are not significant at this sample size, whereas the
pooled difference is significant ($p=0.007$, two-sided Fisher exact test).

\begin{table}[!b]
\caption{Real-robot success under persistent locks, 10 trials per condition.
Locks are software-enforced at the joint-reference layer.}
\label{tab:real_robot}
\centering
\scriptsize
\setlength{\tabcolsep}{6pt}
\resizebox{\columnwidth}{!}{%
\begin{tabular}{lccc}
\toprule
Task & Locked joint & Global-history SAC & CAPABLE \\
\midrule
Open drawer & $j_0$ (seen) &
5/10 (50\%) & \textbf{9/10 (90\%)} \\
Put the object in the drawer & $j_2$ (unseen) &
4/10 (40\%) & \textbf{7/10 (70\%)} \\
Object rearrangement & $j_6$ (seen) &
4/10 (40\%) & \textbf{8/10 (80\%)} \\
\midrule
Overall & -- &
13/30 (43.3\%) & \textbf{24/30 (80.0\%)} \\
\bottomrule
\end{tabular}%
}
\end{table}

\section{Discussion}

A global recurrent encoder can represent fault history, but nothing requires
``$j_0$ locked'' and ``$j_6$ locked'' to occupy related latent regions, and
nothing constrains what it does on a joint it has never seen fail. CAPABLE
constrains that freedom by construction: one temporal function processes every
actuator, so the same learned map interprets a stalled command--response pattern
regardless of joint index. The Jacobian then grounds that shared estimate in
configuration-specific geometry, allowing a transferable representation to
drive a geometrically specific correction.

The ablations support this reading over a purely geometric or purely temporal
alternative. Kinematic information without history reaches 39.4\%, below the
unfactorized baseline, while using separate per-joint history encoders reaches
45.8\%. Full CAPABLE, which combines shared temporal encoding with live
kinematic grounding, reaches 59.3\%.
The mechanism is the conjunction, and the strongest evidence for it is not the
improvement over the frozen VLA but the reproducible advantage over a
parameter-matched global encoder on identical data, across independently 
held-out actuators in the leave-one-actuator-out study.

The cross-fault results delimit where capability of this kind is portable.
Damping, friction, and late-onset locks resemble the trained persistent locks
in how they suppress motion relative to command. Partial effectiveness is
graded rather than near-binary, and range restriction is state-dependent, so
both differ from the training signatures in ways the current formulation does
not capture; extending capability inference to graded and state-dependent
impairments is the natural next step.

\section{Limitations}
\label{sec:limitations}

The broad study remains simulation-heavy, and the hardware evaluation covers
only three task--joint conditions with software-enforced locks. The
leave-one-actuator-out study uses eight rather than all 28 tasks. Training
uses persistent locks only, and CAPABLE does not outperform global-history
SAC under partial effectiveness or range restriction.

\section{Conclusion}
CAPABLE represents actuator faults through remaining physical capability
rather than explicit fault identity. A shared temporal encoder, live
kinematics, and bounded residual control enable transfer to actuators excluded
from fault training while preserving healthy performance. Across 28 LIBERO
tasks, all six leave-one-actuator-out splits, three of five unseen fault
families, and hardware, CAPABLE outperforms a parameter-matched global-history
baseline. These results show that factorized command--response
representations provide a practical route to fault recovery for frozen VLA
policies without fault labels or affected-joint supervision.
\section{Acknowledgment}
AI tools (ChatGPT) was used to generate the line-art robotic arm schematic and to adjust image color balance in Figure~\ref{fig:teaser}. 
\balance

\bibliographystyle{IEEEtran}
\bibliography{IEEEabrv,references}

\end{document}